\documentclass[english]{cccconf}
\usepackage[comma,numbers,square,sort&compress]{natbib}
\usepackage{epstopdf}
\usepackage{amsmath}
\usepackage{bm}
\usepackage{algorithm}
\usepackage{algpseudocode}
\usepackage{amsfonts}
\usepackage{subfigure}
\begin{document}

\title{Optimal Smooth Coverage Trajectory Planning for Quadrotors in Cluttered Environment}

\author{Duanjiao Li\aref{amss},
        Yun Chen\aref{amss},
        Ying Zhang\aref{amss},
        Junwen Yao\aref{amss},
       Dongyue Huang\aref{airs}$^*$,
       Jianguo Zhang\aref{airs},
       Ning Ding\aref{airs}}



\affiliation[amss]{Guangdong Power Grid Co. Ltd., Guangzhou 510220, P.~R.~China
        \email{13922751289@139.com}}
\affiliation[airs]{Shenzhen Institute of Artificial Intelligence and Robotics for Society (AIRS), Shenzhen 518129, P.~R.~China
        \email{huangdongyue@cuhk.edu.cn}}

\maketitle

\begin{abstract}
For typical applications of UAVs in power grid scenarios, we construct the problem as planning UAV trajectories for coverage in cluttered environments. In this paper, we propose an optimal smooth coverage trajectory planning algorithm. The algorithm consists of two stages. In the front-end, a Genetic Algorithm (GA) is employed to solve the Traveling Salesman Problem (TSP) for Points of Interest (POIs), generating an initial sequence of optimized visiting points. In the back-end, the sequence is further optimized by considering trajectory smoothness, time consumption, and obstacle avoidance. This is formulated as a nonlinear least squares problem and solved to produce a smooth coverage trajectory that satisfies these constraints. Numerical simulations validate the effectiveness of the proposed algorithm, ensuring UAVs can smoothly cover all POIs in cluttered environments.
\end{abstract}

\keywords{Traveling Salesman Problem, Genetic Algorithm, Smooth Trajectory Optimization}

\footnotetext{$^*$Corresponding author}
\footnotetext{The work was supported in part by Grant BN00202401031, and in part by the Longgang District Shenzhen's “Ten Action Plan” for Supporting Innovation Projects under Grants LGKCSDPT2024002 and LGKCSDPT2024003.}

\section{Introduction}
\begin{figure}[t!]
    \centering
    \includegraphics[width=0.9\linewidth]{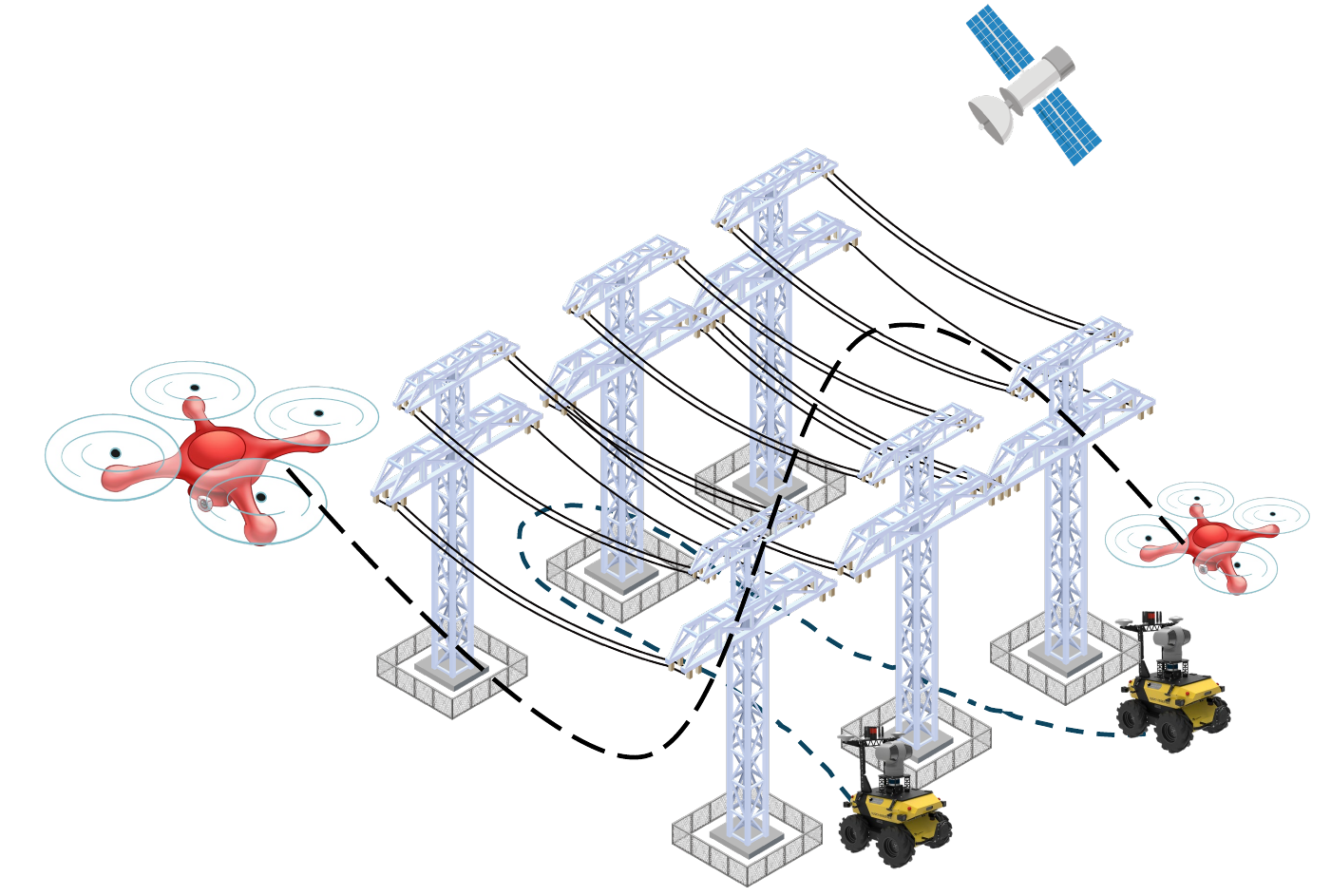}
    \caption{Sketch map of unmanned system operating inspection in the power grid system.}
    \label{fig:front}
\end{figure}

In recent years, with the rapid development of manufacturing industries, unmanned systems have found widespread applications across various fields. Among them, quadrotors have been increasingly utilized in industrial applications such as aerial photography and surveying \cite{huang2024}. As electricity consumption continues to rise, the frequency of power grid maintenance has also increased. Given the high risks and costs associated with manual inspections, the importance of utilizing unmanned systems for autonomous power grid inspections has become increasingly evident \cite{zhou2018}, as shown in Fig~\ref{fig:front}. Substations, as critical components of the power grid system, play an essential role in ensuring seamless inspection across modules within the same facility or between different facilities. The units scheduled for inspection can be abstracted as a series of access points, with drones acting as agents tasked with visiting these points. Consequently, the key focus of this paper is to explore how quadrotors can efficiently cover all inspection points while minimizing energy costs.

In cluttered environments with obstacles, a broad range of motion planning strategies has been extensively studied. Most focus on optimality guarantees and static obstacle avoidance capabilities, such as Expansive Space Trees (EST) \cite{phi2004}, Probabilistic Roadmap Methods (PRM) \cite{Ka1996}, Rapidly exploring Random Trees (RRT) \cite{pha2016}, and optimal RRT (RRT*) \cite{Karaman2010}. Additionally, as noted in \cite{Bou2017}, approaches like multi-RRT* Fixed Node (RRT*FN) and genetic algorithm (GA) have been explored for UAV motion planning in cluttered environments. Yet, these methods fail to adequately address the trajectory generation problem for feasible UAV paths. Moreover, these methods lack real-time applicability and are less suitable for dynamic environments due to high computational costs.
Additionally, some works, such as \cite{Michel2024}, consider not only the shortest path as the sole criterion when addressing the Traveling Salesman Problem (TSP), problem but also incorporate the UAV's dynamic model into the planning process. However, this approach does not take obstacle avoidance into account.

For the trajectory generation of UAVs, several studies have been conducted. For instance, \cite{lai2018_2} generates trajectories with jerk limits, while \cite{lai2018} employs optimal smoothing B-splines and validates the idea through physical experiments. However, these works do not focus on the critical aspect of efficiently accessing predefined trajectory points.

In this paper, we address the defined problem: how to generate smooth and feasible trajectories for UAVs to cover all inspection targets under limited resources. We propose a two-scale solution.
At the first scale, without considering obstacles, all target points are formulated as a TSP, which is an NP-hard problem, solved by a GA.
At the second scale, based on the visitation sequence obtained from the first scale, we use an optimal B-spline to generate trajectories. During this process, obstacles represented by Euclidean Distance Transforms (EDT) are incorporated as constraints to ensure the generated trajectory is feasible for UAV following.

The remaining of the paper is organized as follows: problem
statement is presented in Section II. The proposed algorithms are highlighted in Section III. The numerical simulations are conducted in Section IV. Eventually, some conclusions are remarked in the last section.

\section{Problem Statement}

This paper abstracts the inspection targets in a substation into multiple points, defined as points of interest (POIs), within a 2D cluttered environment containing numerous obstacles. The goal is to generate a smooth trajectory for a UAV to efficiently visit all the inspection points. A quadrotor starts from an initial position, visits all POIs, and then moves to the endpoint along the shortest path while avoiding obstacles. During this process, the UAV navigates along a cost-minimizing path while adhering to dynamic constraints, ensuring the generated trajectory is executable by the UAV.
The proposed algorithm is scalable, capable of handling large and geometrically complex areas. Using sensor-equipped quadrotor to perform coverage tasks presents significant potential for future applications. These flying sensors can serve the same purpose as a network of static sensors, offering rapid, adaptive, and cost-effective solutions for diverse operational scenarios.

\subsection{GA for TSP}

In this problem, we aim to determine the minimum distance path starting from a home base point $p_0$, visiting all points of interest $p_i$ for $i = 1, \ldots, N-1$, each exactly once. The objective is to determine a permutation $\mathcal{S}$ of the sequence:
\begin{equation}
\mathcal{S} = [s_0, s_1, \ldots, s_{N-1}], 
\end{equation}

For each route $s_i \in \mathcal{S}$, the total path length objective function $\mathcal{C}(s)$ is defined, which includes energy and time consumption factors. The total path length is represented as follows:

\begin{equation}\label{cost}
\mathcal{C}(s) = \sum_{i=0}^{N-1} d(s_i, s_{i+1}) E_{\rm f} + d(s_i, s_{i+1}) T_{\rm f}
\end{equation}
where inter-distances $d(s_i, s_j)$ are determined from the set of POIs in the given space. $E_{\rm f}$ and $T_{\rm f}$ are the energy consumption factor and time consumption factor, respectively.

This problem, known as the TSP, is an NP-hard optimization problem, and we propose to use GA to solve this problem and obtain a near-optimal path.
The proposed GA-based solution initializes a population of possible paths and iteratively improves the population by applying selection, crossover, and mutation operations. The optimization is subject to a defined fitness function which aims to minimize the total path length while ensuring that each point is visited only once and the path returns to the end point.

One might use in the genetic algorithm, the selection operation is based on individual fitness (i.e., the inverse of the objective function value). A roulette wheel selection method is used, and the The cumulative probability is defined and determined as follows:

\begin{equation}
\mathbf{P}_i = \sum_{j=1}^{i} \frac{1/\mathcal{C}(s)_i}{\sum_{j=1}^{N} (1/\mathcal{C}(s)_j)}
\end{equation}

\subsection{Optimized B-Spline}
In this problem, we formulate the optimization points as a nonlinear least squares problem. The cost function is defined as a weighted summation of multiple components, including trajectory smoothness $\epsilon_{\rm s}$, obstacle avoidance $\epsilon_{\rm e}$, boundary points $ \epsilon_{\rm b}$, time consumption $\epsilon_{\rm t}$, and POIs $\epsilon_{\rm p}$. 
The parameter $\omega$ with a suffix represents the weight assigned to the corresponding cost function. The goal is to generate a smooth coverage trajectory while minimizing the overall cost, defined as $\epsilon_{\rm total}$. The boundary functions are defined as $g(x)$ and $h(x)$. The mathematical formulation of the cost function is as follows:
\[
\begin{aligned}
    \min_{x} & \quad \frac{1}{2} \| \epsilon_{\rm total} \|_2^2 \\
    \text{s.t.} & \quad g(x) \leq 0, \\
                      & \quad h(x) = 0.
\end{aligned}
\]
where 
\begin{equation}
    \epsilon_{\rm total}  = \omega_{\rm s} \epsilon_{\rm s} + \omega_{\rm b} \epsilon_{\rm b} + \omega_{\rm e} \epsilon_{\rm e} + \omega_{\rm t} \epsilon_{\rm t} + \omega_{\rm p} \epsilon_{\rm p}.
\end{equation}

For the cost function in the objective formulation, the trajectory smoothness is directly evaluated using velocity, acceleration, and jerk as components of the cost function. The time consumption is addressed by introducing a time-scaling variable, which adjusts the time scale as needed.
Additionally, the cost for POIs is directly calculated using the Euclidean distance. For the obstacle cost function, we utilize values derived from the EDT as the specific metric for this cost. A weighting function is applied to regulate its influence within the overall optimization. Additionally, a target safety distance is set to ensure that no safety issues arise when applied to objects of a certain volume.

\subsection{Model of Quadrotors}
\begin{figure}[b]
    \centering
    \includegraphics[width=0.8\linewidth]{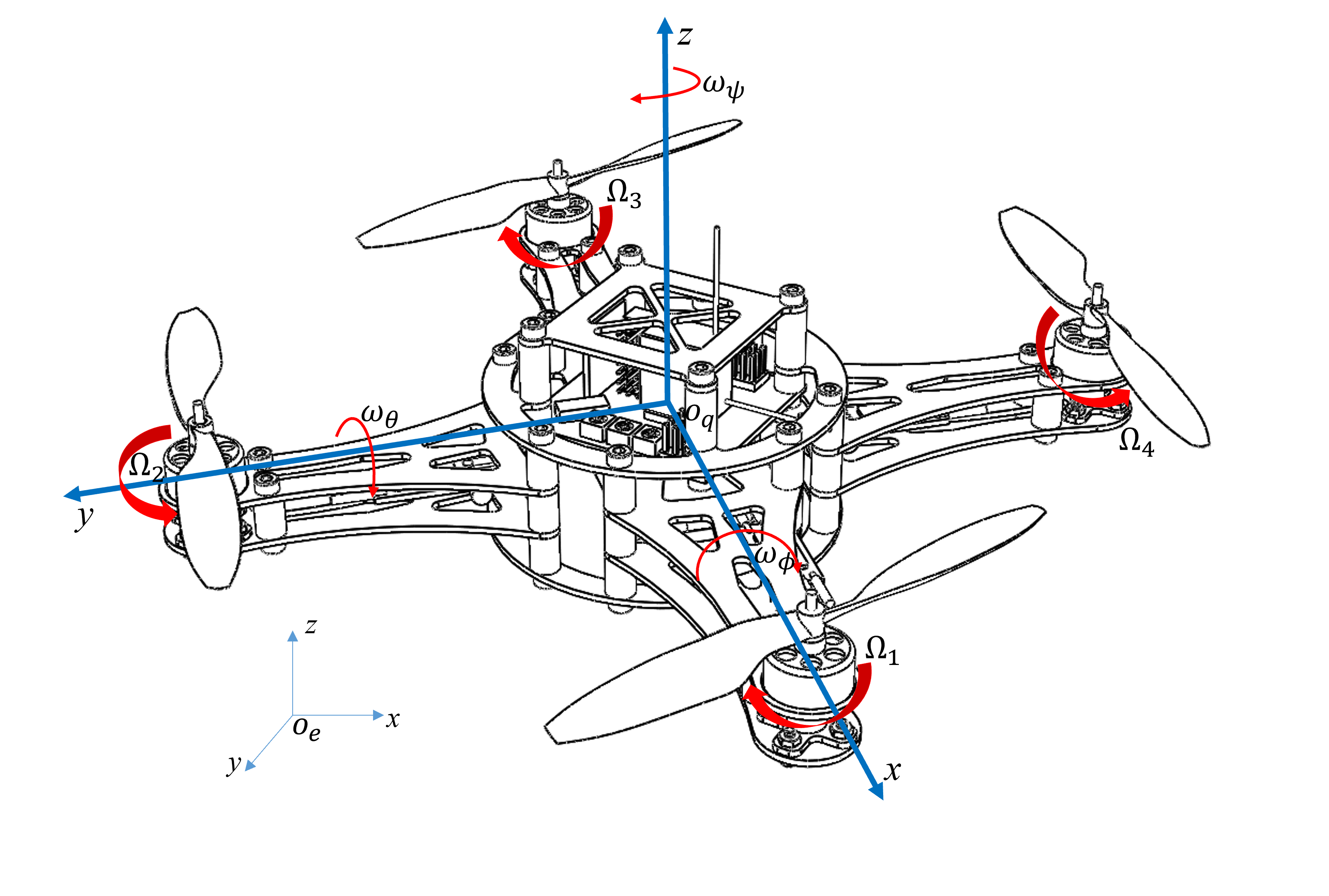}
    \caption{Coordinates of the quadrotors}
    \label{fig:model_frame}
\end{figure}

Before introducing the model of quadrotors, we defined two coordinates, which are body frame, attached at the center of gravity of vehicle and inertial frame, represented by NED (North-East-Down), as shown in Fig.~\ref{fig:model_frame}. The states of the model is described as $\bm{\nu} = [u~v~w~p~q~r]^{\rm{T}}$, represented in the body frame, the pose variable ${\mit{\bm{\eta}}} = [x~y~z~\phi~\theta~\psi]^{\rm{T}}$, both represented in the inertial frame. Defined the transformation matrix as $\bm{J} = {\rm{diag}} ([{\bm{R}}_{\rm{b}}^{\rm{e}},\bm{S}^{{\rm{ - 1}}}])$ \cite{b10}.

As what stated in \cite{huang2024_3}, the candidates UAV model is described as 
\begin{equation}
    \dot {\mit{\bm{\eta}}}  = \bm{J}\bm{\nu},
    \label{eq2}
\end{equation}
\vspace{-17pt}
\begin{equation}
    \bm{M}\dot{\bm{\nu}} + \bm{C}(\bm{\nu})\bm{\nu} + \bm{g} ({\bm{\eta}}) = {\bm{\tau}} +{\bm{\tau}}_{\rm{d}}.
    \label{eq1}
\end{equation}
where $\bm{M}$ is the inertial matrix,  $\bm{C}(\bm{\nu})$ is the Coriolis and centripetal matrix, $\bm{g} ({\bm{\eta}})$ is the restoring force and moment vector, $\bm{\tau}$ is the input force/torque represented under the body frame, and $\bm{\tau}_{\rm d}$ is the disturbance. One notices that the input $\bm{\tau}$ is indirectly affected by rotational speed of the motors.

\section{Algorithms Description}

\begin{figure}[h]
    \centering
    \includegraphics[width=0.97\linewidth]{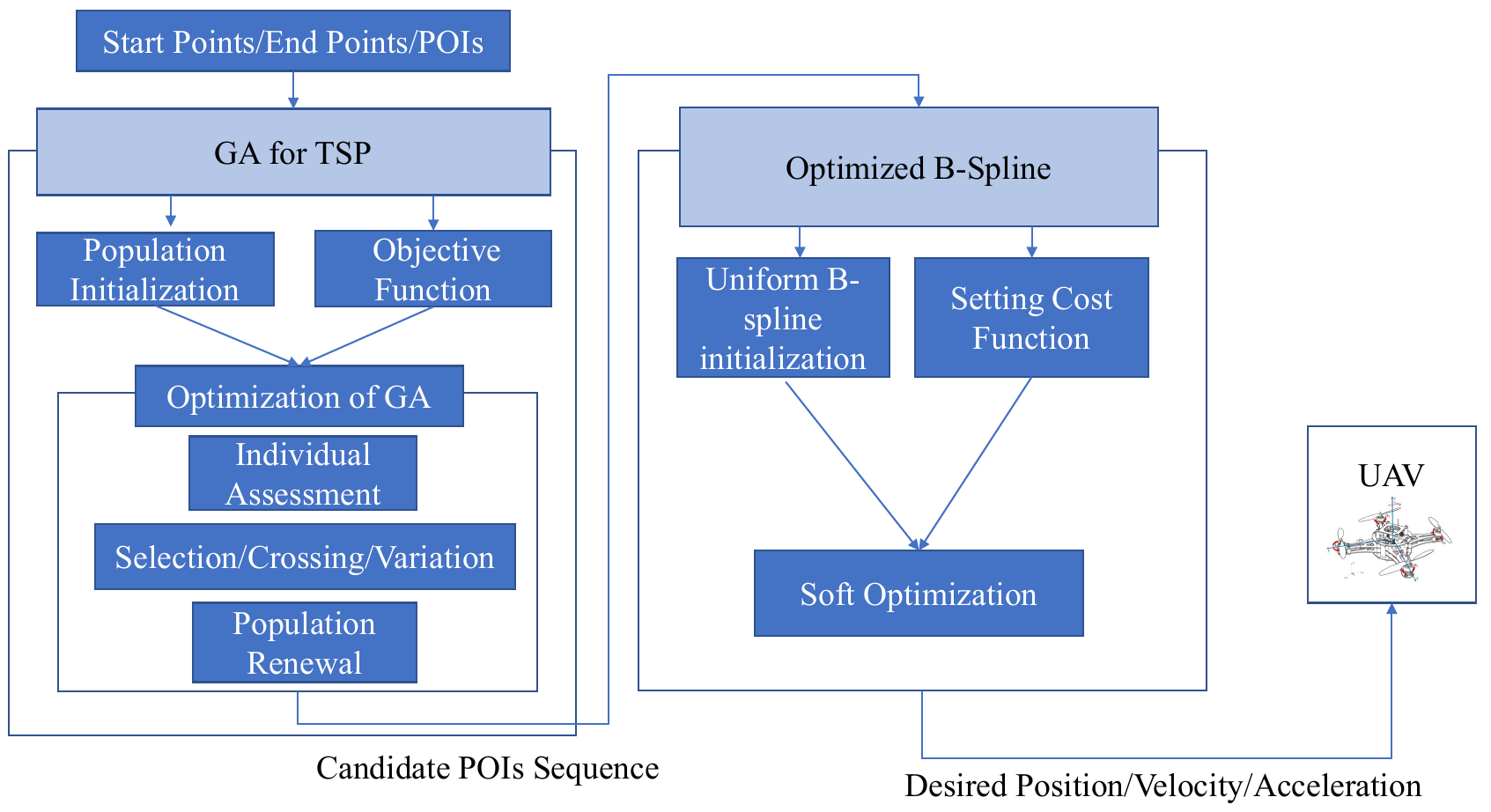}
    \caption{Overview of the proposed planning algorithm.}
    \label{fig:algo}
\end{figure}

In this section, we mainly introduce the algorithm used in planning the optimal smooth coverage trajectory, as shown in Fig~\ref{fig:algo}. Firstly, we give a detailed description on the GA for optimizing the TSP. Then a detailed optimized b-spline algorithm is introduced. 
\subsection{GA for TSP Optimization}

The proposed algorithm employs a GA to solve the TSP, aiming to minimize the total path cost, which incorporates both energy and time consumption. The algorithm starts by initializing a population \( \mathcal{P} \), consisting of random permutations of POIs excluding the fixed start and end points. The initial best sequence of the traveling POIs and cost are computed using an objective function, which calculates the energy consumption and travel time for each segment of the route. The optimization process runs for \( \mathcal{G}_{\rm max} \) generations, where in each generation, the individuals are evaluated to determine their fitness based on the objective function. The best solution is updated if a better sequence is found. The `PopulationEvoFunc' function is then used to evolve the population by selecting the top-performing individuals for crossover and mutation, resulting in offspring paths that are fixed by the `RepairPath' function to ensure validity, i.e., that each POI is visited exactly once.

The cost function \( \mathcal{C} \), as defined in (\ref{cost}), involves calculating the energy consumption and time required for traversal between POIs, summed over all consecutive POIs pairs along the path. During the evolution phase, crossover operations are performed by selecting crossover points between two individuals to produce offspring, which are then repaired to remove any duplicated POI while maintaining the fixed start and end points. The algorithm terminates after \( \mathcal{G}_{\rm max} \) generations, returning the best path found and its associated cost, thus providing an optimized sequence of POIs, denoted as \(\mathcal{S}^*\), visits that considers both energy and time efficiency.

\begin{algorithm}[!t]
\caption{Genetic Algorithm}
\begin{algorithmic}[1]
\State \textbf{Input:} $N$, $d$, $\rm{size}(\mathcal{P})$, $\mathcal{G}_{\rm max}$, $E_{\rm f}$, $T_{\rm f}$
\State \textbf{Output:} $\mathcal{S}^*$, $\mathcal{C}_{\rm min}$

\State \textbf{Initialization:} $\mathcal{P} \in \mathbb{R}^{{\rm{size}(\mathcal{P})} \times (N-2)}$ for each individual (excluding start and end)

\State $\mathcal{S}^* \gets [1, \mathcal{P}(1,:), N]$
\State $\mathcal{C}_{\rm min} \gets \texttt{ObjectiveFunc}(\mathcal{S}^* , d, E_{\rm f}, T_{\rm f})$

\For {$\mathcal{G}_{\rm max}$}
    \For {$i = 1 \to \rm{size}(\mathcal{P})$}
        \State $\mathcal{S} \gets [1, \mathcal{P}(i,:), N]$
        \State $\mathcal{C}(i) \gets \texttt{ObjectiveFunc}(\mathcal{S}, d, E_{\rm f}, T_{\rm f})$
    \EndFor
    
    \State $\texttt{currentMinCost}, \texttt{idx} \gets \min(\mathcal{C})$
    \If {$\texttt{currentMinCost} < \mathcal{C}_{\rm min}$}
        \State $\mathcal{C}_{\rm min} \gets \texttt{currentMinCost}$
        \State $\mathcal{S}^* \gets [1, \mathcal{P}(\texttt{idx}, :), N]$
    \EndIf

    \State $\mathcal{P} \gets \texttt{PopulationEvoFunc}(\mathcal{P}, \mathcal{C}, N)$
\EndFor

\State \hrulefill

\Function{PopulationEvoFunc}{$\mathcal{P}, \mathcal{C}, N$}
    \State $\mathcal{P}_{\rm s} \gets \text{top 50\% individuals}$ based on $\text{costs}$
    \State $\mathcal{P}_{\rm n} \gets \mathcal{P}_{\rm s}$
    \While {$|\mathcal{P}_{\rm n}| < |\mathcal{P}|$}
        \State $i, j \gets \text{random selection from } \text{selected}$
        \If {$i \neq j$}
            \State $\texttt{Prt}_{\rm c} \gets \text{random index}$
            \State $\texttt{RPath1} \gets [\mathcal{P}_{\rm s}(i, 1:\text{Prt}_{\rm c} ), \mathcal{P}_{\rm s}(j, \texttt{Prt}_{\rm c} +1:\text{end})]$
            \State $\texttt{RPath2} \gets [\mathcal{P}_{\rm s}(j, 1:\texttt{Prt}_{\rm c} ), \mathcal{P}_{\rm s}(i, \texttt{Prt}_{\rm c} +1:\text{end})]$
            \State $\texttt{RPath1} \gets \texttt{RepairPath}
            (\texttt{RPath1}, N-2)$
            \State $\texttt{RPath2} \gets \texttt{RepairPath}
            (\texttt{RPath2}, N-2)$
            \State $\mathcal{P}_{\rm n} \gets \text{append}\ \ \texttt{RPath1, RPath2} $
        \EndIf
    \EndWhile
    \State \textbf{Return:} $\mathcal{P}_{\rm n}$
\EndFunction

\State \hrulefill

\Function{ObjectiveFunc}{$\mathcal{S}, d, E_{\rm f}, T_{\rm f}$}
    \State $E \gets 0$, $T \gets 0$
    \For {$i = 1 \to \text{size($\mathcal{S}$)} - 1$}
        \State $E \gets E + E_{\rm f} \cdot d(\mathcal{S}(i), \mathcal{S}(i+1)) $
        \State $T \gets T + T_{\rm f} \cdot d(\mathcal{S}(i), \mathcal{S}(i+1))$
    \EndFor
    \State $\mathcal{C} \gets E + T$
    \State \textbf{Return:} $\mathcal{C}$
\EndFunction

\State \hrulefill

\Function{RepairPath}{$\texttt{path}$, $N$}
    \State $\texttt{RPath} \gets \texttt{path}$
    \For {$i = 1 \to \text{size} \texttt{(path)}$}
        \If {$\sum(\texttt{RPath} == \texttt{RPath}(i)) > 1$}
            \State $\texttt{unusedPrt} \gets \{2, \dots, N+1\} \setminus \texttt{RPath}$ 
            \State $\texttt{RPath}(i) \gets \texttt{unusedPrt}(1)$
        \EndIf
    \EndFor
    \State \textbf{Return:} $\texttt{RPath}$
\EndFunction

\end{algorithmic}
\end{algorithm}

\subsection{Optimized B-Spline}

In this subsection, we start by initializing a binary grid (\texttt{bw}) to represent the environment. Each obstacle (\texttt{obs}) is defined by its coordinates, and the corresponding grid index is computed using the \texttt{pos2grid} function. It converts coordinates $\texttt{(x,y)}$ into grid indices $\texttt{(idx,idy)}$ by dividing by step, rounding. For each obstacle, we mark the cells within its defined radius ($r_{\rm obs}$) as occupied in the binary map. The resolution of the grid is determined by a step size ($\lambda$), which influences the granularity of obstacle representation. Finally, we compute the \texttt{EDT} of the grid, which provides a distance map indicating the proximity of each unoccupied cell to the nearest obstacle, giving a continuous representation of obstacle influence in the environment.

\begin{algorithm}[!h]
\caption{Obstacle Map}
\begin{algorithmic}[1]
\State \textbf{Input:} \texttt{obs}, $\lambda$, $r_{\rm obs}$
\State \textbf{Output:} \texttt{EDT}

\State Initialize binary map $\texttt{bw}$ of the environment.
\State Set step $\lambda$ and obstacle radius $r_{\rm obs}$.

\For{each obstacle in \texttt{obs}}
    \State Calculate grid index for obstacle position: $(\texttt{idx}, \texttt{idy}) \gets \texttt{pos2grid}(\texttt{obs})$
    
    \For{each \texttt{x\_offset} and \texttt{y\_offset} within \texttt{obs\_radius}}
        \If{distance is within \texttt{obs\_radius}}
            \State Calculate new indices in grid: $(\texttt{idx}, \texttt{idy})$
            \If{\texttt{idx}, \texttt{idy} are within the bounds of $\texttt{bw}$}
                \State Set $\texttt{bw}[\texttt{idx}, \texttt{idy}] = 1$ 
            \EndIf
        \EndIf
    \EndFor
\EndFor

\State $\texttt{EDT} \gets \lambda \cdot \texttt{bwdist(bw, 'euclidean')} $
\State \textbf{Return:} \texttt{EDT}
\end{algorithmic}
\end{algorithm}

Then, we construct a cost function based on the established EDT obstacle map, and incorporate it into the B-spline optimization process for trajectory planning. Additionally, we consider other costs such as trajectory smoothness, as described in (1). Furthermore, the sequence of visiting points obtained from the GA is also incorporated as a cost term. The overall goal of this process is to solve for the optimal control points for the trajectory.

\section{Numerical Simulations}
In the simulation, we applied our algorithm in a quadrotor in a clutter environment with traveling all POIs, with specific parameters listed in Table \ref{UAVPar}. As the proposed planning algorithm is 2D and depending on mapping, assuming that all the positions of the collisions are pre-known, and all the POIs are located at the height of 5~m.

\begin{figure}[h]
        \centering
		\subfigure[\label{fig:map1}] 
{\includegraphics[width = 0.2\textwidth]{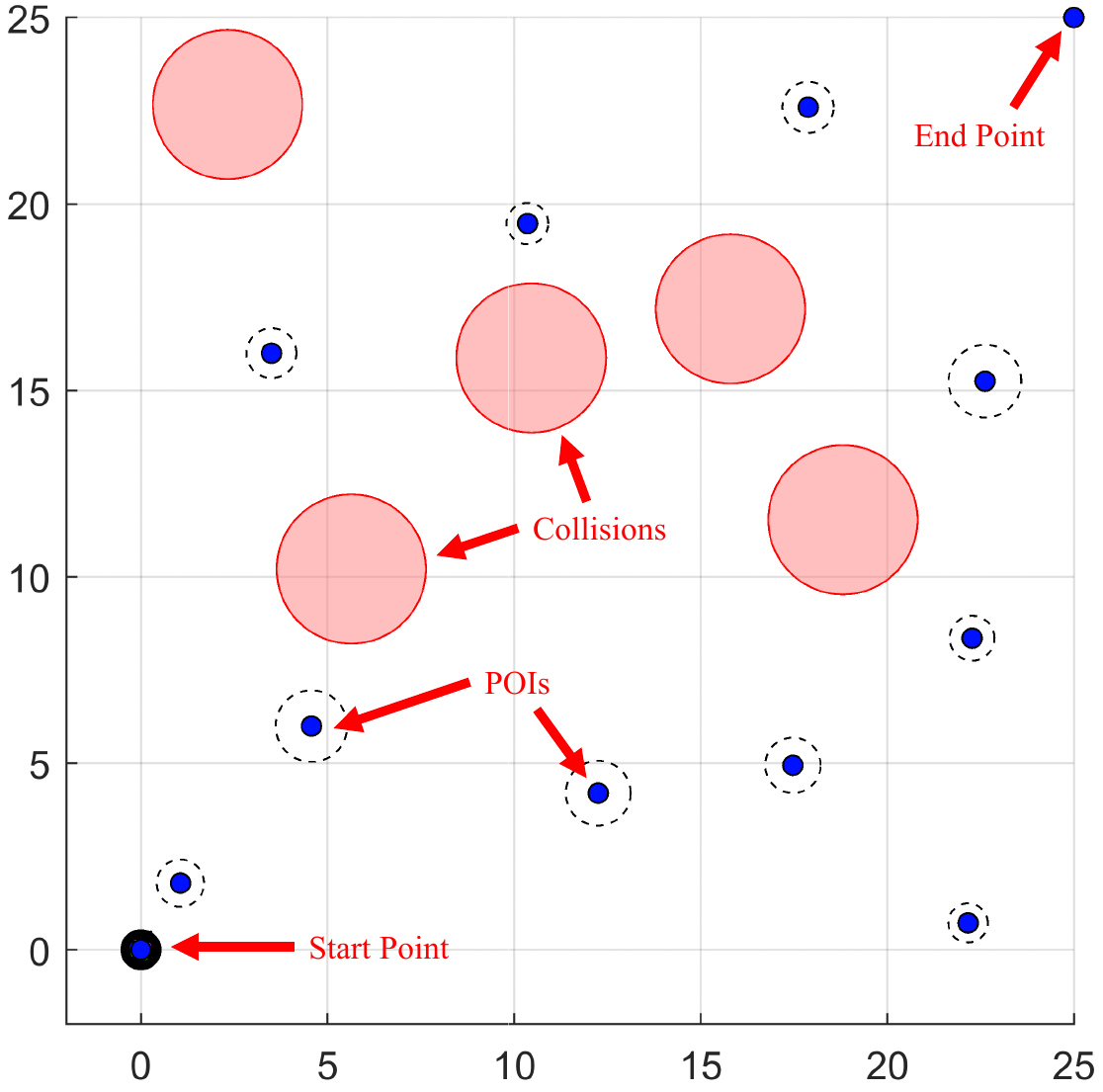}}~
		\subfigure[\label{fig:map2}]
{\includegraphics[width = 0.23\textwidth]{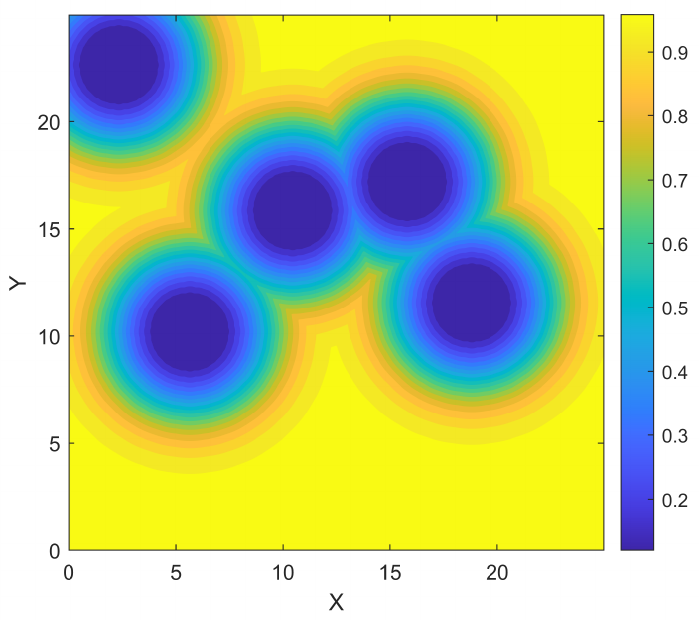}}
\caption{Maps for operating. (a) The top view of the map for operating. (b) Cost map generated through EDT. }
\label{fig:map}
\end{figure}

The scenario in this numerical simulation is depicted in Fig~\ref{fig:map}. Specifically, as shown in Fig.~\ref{fig:map1}, red areas represent obstacles, while the blue points enclosed by dashed lines indicate the POIs. The dashed lines denote the coverage range, simulating the maximum field of view of UAV. The starting and ending points are also marked. The UAV begins at the starting point, visits each POI while avoiding obstacles, and ultimately reaches the endpoint to complete the inspection task. Additionally, the cost map generated using the EDT is shown in Fig.~\ref{fig:map2}.

\begin{table}[!t]
  \centering
  \caption{UAV Parameters}
  \label{UAVPar}
  \begin{tabular}{l|l}
    \hhline
    Mass (kg) & 0.5  \\ \hline
        Moment of Inertia  ($\text{kg} \cdot \text{m}^2$)  & Diag(2.93,2.32,4) $\times 10^{-3}$ \\ \hline
        Wheelbase (m) & 0.3  \\ \hline
        Thrust Coefficient ($\text{N}/\text{rpm}^2$) & 6.11 $\times 10^{-8}$  \\ \hline
        Moment Coefficient ($\text{Nm}/\text{rpm}^2$)& 1.50 $\times 10^{-9}$  \\
    \hhline
  \end{tabular}
\end{table}

\begin{figure}[h]
    \centering
    \includegraphics[width=0.95\linewidth]{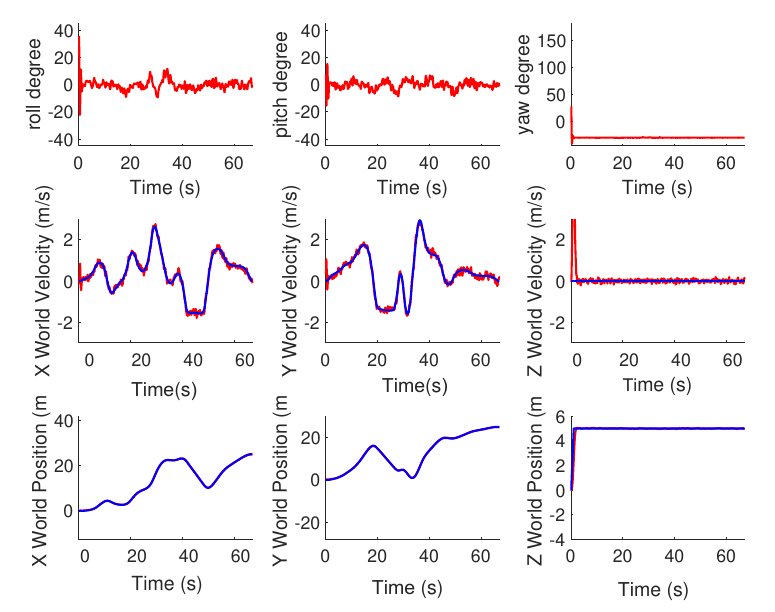}
    \caption{Numerical simulation results of the state variables during the simulation.}
    \label{fig:states}
\end{figure}

We apply the proposed algorithm to the UAV and scenario described above, and the resulting numerical simulations are shown below. As depicted in Fig.~\ref{fig:sim_1}, the green line represents the smooth coverage trajectory planned by the algorithm, while the blue line illustrates the actual trajectory executed by the UAV. Details of each state of the UAV are presented in the Fig.~\ref{fig:states}. The first row shows the roll, pitch, and yaw angles over time. The second row illustrates the velocities. The third row depicts the position, respectively. The blue line represents the reference trajectory from the planner, while the red line indicates the UAV trajectory execution.

\begin{figure}[h]
    \centering
    \includegraphics[width=0.88\linewidth]{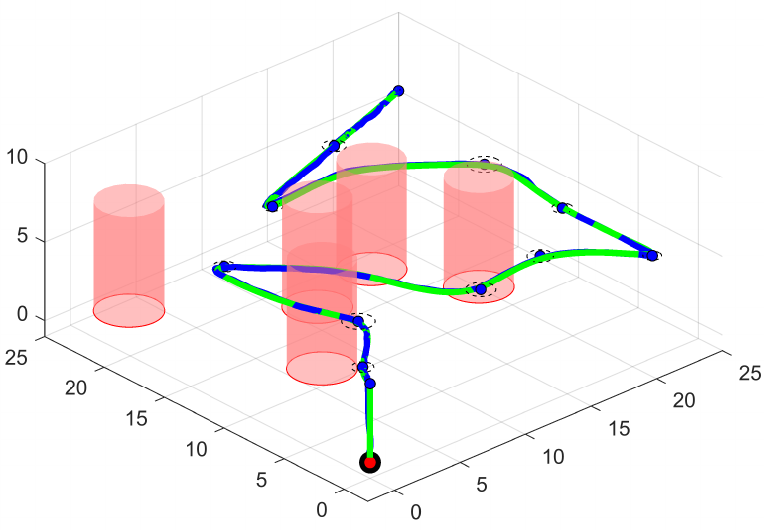}
    \caption{The trajectory generated in the full cycle numerical simulation. The green line denotes the planned trajectory. The blue line denotes the estimated UAV trajectory.}
    \label{fig:sim_1}
\end{figure}

\begin{figure}[h]
        \centering
		\subfigure[\label{fig:sim1}] 
{\includegraphics[width = 0.23\textwidth]{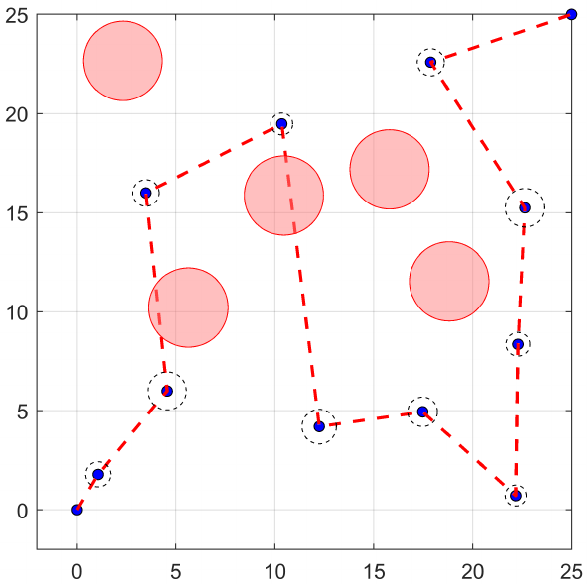}}~
		\subfigure[\label{fig:sim2}]
{\includegraphics[width = 0.23\textwidth]{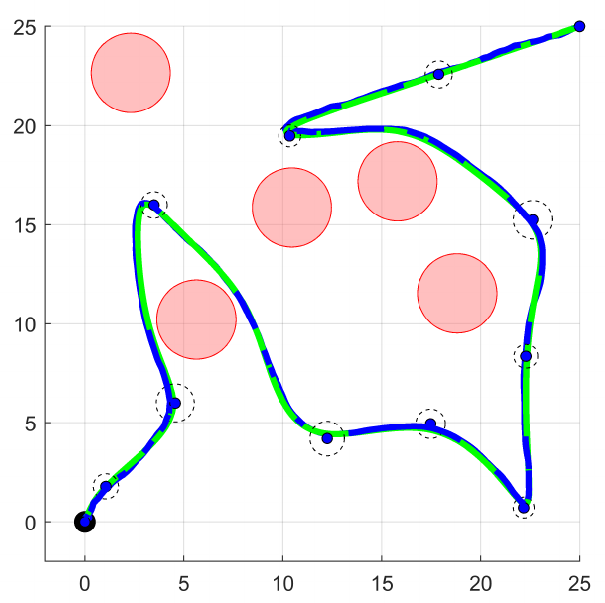}}
\caption{Results of the simulations. (a) The top view of the simulation after the optimization of GA. (b) The top view of the simulation after the optimization of b-spline. }
\label{fig:sim}
\end{figure}

In Fig.~\ref{fig:sim1}, the POIs are first processed through the GA to determine an initial sequence of visitations, represented by the red dashed line. Since obstacle constraints are ignored during this optimization, the resulting sequence serves as a preliminary solution. Next, the GA-optimized sequence is fed into the optimized B-spline algorithm, producing a smooth trajectory that the UAV can feasibly follow, as shown in Fig.~\ref{fig:sim2}. It is noticeable that the POIs visitation order changes after incorporating the optimized B-spline. This change arises from the need to account for obstacle avoidance, minimize traversal costs, and ensure smooth UAV operation, leading to adjustments in the cost function.
It is also reflected in the detailed state variations. The UAV avoids large angular maneuvers, which not only contributes to energy savings but also aids in maintaining alignment with inspection targets during operation. These advantages are particularly significant in real-world industrial scenarios, ensuring higher inspection quality and operational efficiency.

The results from Fig.~\ref{fig:states} demonstrate the effectiveness of the proposed algorithm in generating smooth and collision-free trajectories for UAV operations in cluttered environments. The roll and pitch angles exhibit minor oscillations, reflecting robust stabilization, while the yaw remains stable, indicating consistent heading control. The X and Y velocities show periodic variations aligned with acceleration and deceleration phases, ensuring energy-efficient motion between POIs. The positional trajectories in X and Y directions are smooth and closely follow the planned path, confirming high tracking accuracy. Additionally, the constant Z position highlights reliable altitude maintenance, essential for stable sensor operations. Overall, the control system handles disturbances effectively, ensuring real-time applicability and energy-efficient performance during inspection tasks.

\section{Conclusion}

In this paper, we have proposed an optimization planning algorithm for UAVs to generate smooth and coverage trajectories for visiting a series of POIs in cluttered environments. Specifically, we have first employed a GA to solve the TSP for candidate POIs and subsequently optimized the results using an optimal B-spline, incorporating cost functions such as smoothness and obstacle avoidance. Through the application of the proposed algorithm in UAV simulations, we have demonstrated its effectiveness. The simulation results have shown that during the full cycle, the UAV has not exhibited any significant angular maneuvers, with the maximum attitude angle variation remaining below 20 degrees. Additionally, the velocity profile has been smooth without any step-like changes. These simulation outcomes have proven beneficial for the application of this algorithm in UAV inspection tasks, laying a solid foundation for future real-world deployments.

\end{document}